\documentclass{Interspeech}

\usepackage{algorithm}
\usepackage{tcolorbox}
\usepackage{algpseudocode} 
\usepackage{pgfplots}
\usepackage{pgfplotstable} % For table-based plotting
\usetikzlibrary{patterns}
\pgfplotsset{compat=1.17}
\usepackage{multirow}
\usepackage{cite}
\interspeechcameraready

\title{Contextual Paralinguistic Data Creation for  Multi-Modal Speech-LLM: Data Condensation and Spoken QA Generation}
\author[affiliation={1}]{Qiongqiong}{Wang}
\author[affiliation={1}]{Hardik B.}{Sailor}
\author[affiliation={1}]{Tianchi}{Liu}
\author[affiliation={1}]{Ai Ti}{Aw}

\affiliation{Institute for Infocomm Research (I$^2$R)}{Agency for Science, Technology and Research (A$\star$STAR)}{Singapore}
\email{\{wang\underline{\enskip}qiongqiong, sailor\underline{\enskip}hardik\underline{\enskip}bhupendra, liu\underline{\enskip}tianchi, aaiti\}@i2r.a-star.edu.sg}

\usepackage{comment}

\begin{document}

\maketitle

\begin{abstract}
Current speech-LLMs exhibit limited capability in contextual reasoning alongside paralinguistic understanding, primarily due to the lack of Question-Answer (QA) datasets that cover both aspects. We propose a novel framework for dataset generation from in-the-wild speech data, that integrates contextual reasoning with paralinguistic information. It consists of a pseudo paralinguistic label-based data condensation of in-the-wild speech and LLM-based Contextual Paralinguistic QA (CPQA) generation. The effectiveness is validated by a strong correlation in evaluations of the Qwen2-Audio-7B-Instruct model on a dataset created by our framework and human-generated CPQA dataset. The results also reveal the speech-LLM's limitations in handling empathetic reasoning tasks, highlighting the need for such datasets and more robust models. The proposed framework is first of its kind and has potential in training more robust speech-LLMs with paralinguistic reasoning capabilities.
\end{abstract}
% %For validation, we evaluate the Qwen2-Audio-7B-Instruct model on a dataset created by our proposed framework. The results reveal the speech-LLM's limitations in handling empathetic reasoning tasks, highlighting the need such datasets and more robust models. The strong correlation with evaluations on human-generated CPQA data further validates our framework's effectiveness for generating such datasets.
\section{Introduction}
    Rapid advancements in large language models (LLMs) have sparked significant interest in multimodal models that integrate LLMs with speech modalities. Recent speech-LLMs, such as GPT-4~\cite{gpt4}, Qwen-audio~\cite{qwen,qwen2}, SALMONN~\cite{salmonn}, and MERaLiON-AudioLLM~\cite{meralion, huzaifah2024towards}, have demonstrated remarkable performance in handling speech-based tasks. 
    Some speech-LLMs, in particular, focus on contextual reasoning properties derived from speech~\cite{wang24blsp, Rubenstein23,Chen23,Lin24, Wang2023BLSPBL}. 
    
    Several studies have attempted to train models to understand emotions in speech and respond empathetically~\cite{Lin24,wang24blsp, kimparalinguistics, kang2024frozen}. Notable approaches are presented in \cite{wang24blsp} and \cite{kang2024frozen}, where the authors introduced training strategies to enhance QA performance by incorporating paralinguistic information from existing speech emotion datasets. These models, however, exhibit limited capabilities in contextual reasoning alongside paralinguistic understanding, primarily due to the lack of QA datasets that cover both aspects. To incorporate paralinguistic cues, QA generation must extend beyond linguistic features. We refer to such QA as contextual paralinguistic QA (CPQA).

Creating contextual paralinguistic data presents two major challenges and significantly hinders progress in this area. First, the availability of relevant metadata is limited. Publicly accessible speech datasets with paralinguistic labels are typically small and task-specific. Emotion-labeled data is even rarer than attributes like speaker identities or gender. While vast amounts of speech data exist, obtaining well-annotated data with reliable paralinguistic metadata remains difficult. Unlike textual data, speech annotation is more complex, as it requires both transcriptions and precise labeling of paralinguistic features. Some paralinguistic labels, such as speaker or gender, are relatively objective, while others, such as emotion, are particularly challenging due to their subjective nature.  Emotion labeling requires multiple annotators and majority voting, with low-agreement samples often discarded.

Second, CPQA generation is non-trivial. High-quality QA requires that questions comprehensively cover all relevant aspects of performance and that the reference answers are both accurate and unbiased. Existing benchmarks, such as AudioBench~\cite{audiobench} and Dynamic-Superb~\cite{dynamicsuperb}, AIR-Bench \cite{AIRbench}, OpenASQA \cite{OpenASQA} and MMAU \cite{MMAU} evaluate not only speech understanding but also paralinguistic tasks, including emotion recognition. However, the QA in these benchmarks primarily derives from speech emotion datasets, such as IEMOCAP~\cite{iemocap} and MELD~\cite{meld}, which focus on isolated emotion-related tasks. These QA datasets typically frame QA pairs in a direct manner (e.g., explicitly asking for an emotion label) without incorporating contextual reasoning.

To address these challenges, we propose a novel framework for dataset generation from in-the-wild speech data. The framework consists of data condensation and automated CPQA generation, specifically focusing on emotion aspect. To the best of our knowledge, this work is the first to integrate contextual reasoning with paralinguistic cues. The proposed framework supports both training and evaluation data creation for speech-LLMs. In this paper, we validate the framework by generating an evaluation dataset and comparing it to human-generated QA sets in LLM evaluations. Our contributions are as follows:
\begin{itemize}
    \item Speech Data Condensation: We address the challenge of limited and noisy paralinguistic labels by integrating categorical and dimensional emotion recognition models, resulting in more accurate pseudo emotion labels.
    \item CPQA Generation via LLM: We automate the generation of QA pairs using LLMs to capture both paralinguistic cues and contextual reasoning. These generated pairs are compared with human-generated sets in speech-LLM evaluations. This approach has the potential to generate large-scale training datasets for speech-LLMs. 
    \item Open-Source Evaluation Dataset: We provide an open-source benchmark comprising 480 speech samples paired with CPQA pairs.\footnote{https://huggingface.co/datasets/MERaLiON/CPQA-Evaluation-Set}
\end{itemize}
%\begin{itemize}
%    \item Speech Data Condensation: We introduce a pipeline for automatic generation of pseudo emotion labels by integrating both categorical and dimensional speech emotion recognition (SER) models for improved labeling accuracy. %It can be extended to gather fine-tuning data. %Additionally, gender labels are incorporated using a gender recognition model.
%    \item CPQA Pair Generation via LLM: We propose a method for automated generating QA pairs using LLMs, covering both paralinguistic related and contextual reasoning questions, and provide a comparison with human-generated QA sets in LLM evaluations.
%    \item Open-Source Evaluation Dataset: We release an evaluation dataset of 480 speech samples with 3,099 LLM-generated CPQA pairs, providing a benchmark evaluation. \textcolor{red}{The link will be released upon paper acceptance.}
    %\item Benchmark for Speech-LLM's Capabilities in Paralinguistic Contextual Reasoning: We provide an evaluation to assess the paralinguistic understanding of speech-LLM\textcolor{red}{s?}, highlighting the strengths and limitations of current models  in recognizing emotional, gender, etc. as well as complex reasoning questions. \textcolor{red}{Should we keep this item? We only evaluated one LLM.}
%\end{itemize}

\section{Proposed data generation framework}
\begin{figure}[t]
  \centering
  \includegraphics[width=0.9\linewidth]{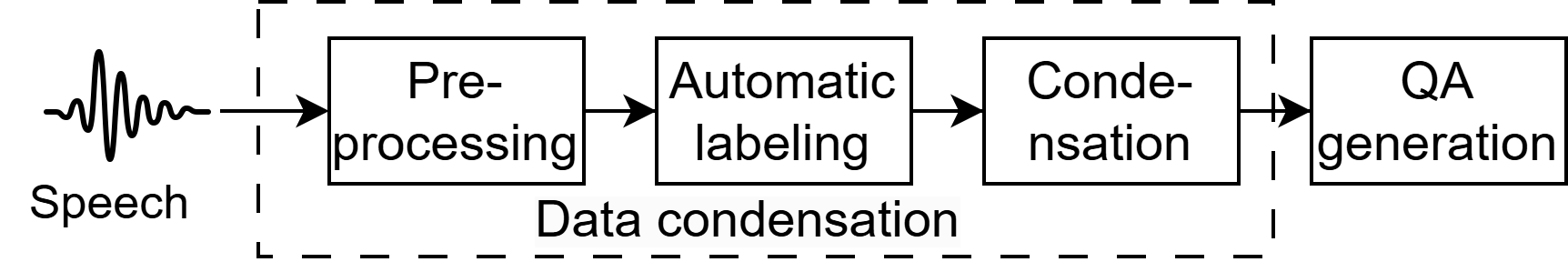}
  \caption{Diagram of dataset creation framework.}
  \label{fig:Diagram}
\end{figure}

We propose a novel framework for CPQA dataset generation from in-the-wild speech data. The framework consists of data condensation and automated CPQA generation (see Figure~\ref{fig:Diagram}).

\subsection{Data condensation pipeline}
Emotion annotation is challenging due to its subjective nature~\cite{Ortony1990,Plutchik2013}, requiring multiple annotators and majority voting for reliability, which makes it resource-intensive. Additionally, most in-the-wild speech data predominantly exhibit a neutral emotional tone. An unbalanced distribution skewed toward the ``neutral'' category would limit the development of an effective dataset for empathetic speech-LLMs, as it would fail to provide meaningful insights for emotion-related tasks.

To address these challenges, we propose a data condensation pipeline. First, we employ automatic speech emotion recognition (SER) tools to avoid the high cost of human annotation. The limitations of these tools in SER accuracy of spontaneous speech~\cite{emobox} are mitigated through a condensation technique. This technique filters out low-confidence samples and ensures a balanced distribution across emotion categories, which is crucial for maintaining fairness, robustness, and generalizability within the dataset.

\subsubsection{Pre-processing}
\label{sssec:pre-processing}
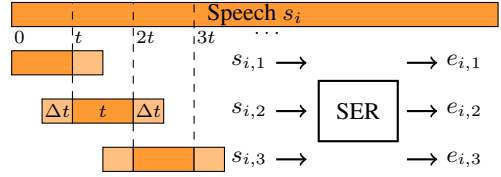
\begin{figure}[t]
    \centering
    \begin{tikzpicture}[scale=0.8]
    \begin{scope}[every node/.style={font=\scriptsize}] % Apply small font size
        % Full audio bar
        \draw[fill=orange!80] (0,2.4) rectangle (8,2.8) node[midway] {\small Speech $s_i$};

        % Dashed segmentation lines
        \draw[dashed] (1,1.2) -- (1,2.8);
        \draw[dashed] (2,0.4) -- (2,2.8);
        \draw[dashed] (3, 0) -- (3,2.8);
        
        % Time Scale Labels
        \node[below] at (0.13,2.46) {$0$};
        \node[below] at (1.1,2.46) {$t$};
        \node[below] at (2.2,2.46) {$2t$};
        \node[below] at (3.2,2.46) {$3t$};
        \node[below] at (4.2,2.46) {$\ldots$};
        
        % Overlapping segments on separate lines
        % First segment
        \draw[fill=orange!80] (0,1.6) rectangle (1,2);
        \draw[fill=orange!50] (1,1.6) rectangle (1.5,2);

        % Second segment
        \draw[fill=orange!50] (0.5,0.8) rectangle (1,1.2) node[midway] {\scriptsize $\Delta t$};
        \draw[fill=orange!80] (1,0.8) rectangle (2,1.2) node[midway] {\scriptsize $t$};
        \draw[fill=orange!50] (2,0.8) rectangle (2.5,1.2) node[midway] {\scriptsize $\Delta t$};

        % Third segment
        \draw[fill=orange!50] (1.5,0) rectangle (2,0.4);
        \draw[fill=orange!80] (2,0) rectangle (3,0.4);
        \draw[fill=orange!50] (3,0) rectangle (3.5,0.4);

        % Arrows to SER
        \draw[->, thick] (4.35,1.8) -- (4.85, 1.8) node[midway, left, xshift=-2mm] {\small $s_{i,1}$};
        \draw[->, thick] (4.35,1) -- (4.85,1) node[midway, left, xshift=-2mm] {\small $s_{i,2}$};
        \draw[->, thick] (4.35,0.2) -- (4.85,0.2) node[midway, left, xshift=-2mm] {\small $s_{i,3}$};
        % SER Box
        \draw[fill=white, draw=black, thick] (5.05,0.5) rectangle (6.35,1.5) node[midway] {\small SER};

        % Arrows from SER to output
        \draw[->, thick] (6.5,1.8) -- (7, 1.8) node[midway, right, xshift=2mm] {\small $e_{i,1}$};
        \draw[->, thick] (6.5,1) -- (7,1) node[midway, right, xshift=2mm] {\small $e_{i,2}$};
        \draw[->, thick] (6.5,0.2) -- (7,0.2) node[midway, right, xshift=2mm] {\small $e_{i,3}$};
    \end{scope} 
    \end{tikzpicture}
    \caption{Illustration of speech segmentation. SER is performed on segments $s_{i,j}$ of length $t+2\Delta t$  and creates emotion labels $e_{i,j}$ for segments of length $t$.}
    \label{fig:segmentation}
\end{figure}

Speech samples often contain silence or noise and vary in length, requiring pre-processing for SER. We use voice activity detection to remove non-speech, retaining segments $
S = \{s_1, s_2, \dots, s_N\}$.
Each $s_i$ is then segmented into sub-segments $s_i = \{s_{i,1}, s_{i,2}, \dots, s_{i,M_i}\}$ of $t+2\Delta t$ windows with $2\Delta t$ overlaps. As shown in Figure~\ref{fig:segmentation}, SER is applied on $s_{i,j}$ and assign an emotion label $e_{i,j}$ to the middle $t$ of $s_{i,j}$, considering the contexts of $\Delta t$ before and afterward. 
\subsubsection{Automatic emotion label annotation}
Emotion recognition follows two paradigms: discrete emotion categories (e.g., happy, angry, sad, neutral) and  dimensional representations(valence, arousal, dominance)~\cite{russell1977}. %typically valence (pleasantness of the stimulus), arousal (intensity of emotion provoked by the stimulus) and dominance (degree of control exerted by the stimulus).
Valence is closely related to sentiment. while discrete emotions are intuitive, they struggle with mixed emotions and data scarcity for certain classes. Dimensional SER models provide a broader view but lack of interpretability. A hybrid approach combining both paradigms improves SER performance by improving prediction confidence and enabling inference beyond predefined categories, resulting in a more flexible and robust system. For simplicity, we focus on the valence and, inspired by the sentiment dictionary~\cite{Demszky2020}, define a mapping rule between sentiment classes and the emotion categories based on valence values.
%\begin{itemize}
%    \item Valence: the pleasantness of the stimulus
%    \item Arousal: the intensity of emotion provoked by the stimulus
%    \item Dominance: the degree of control exerted by the stimulus
%\end{itemize}

%\begin{table}[h]
%    \centering
%    \begin{tabular}{lcc}  % 'l' for left-align, 'c' for center-align
%        \toprule
%        Model & Accuracy & F1 score \\ 
%        \midrule
%        base_finetuned  &36.15&37.01  \\ 
%        plus_seed       &48.89&48.19 \\ 
%        plus_base       &48.59&46.65 \\ 
%        plus_large      &43.50&44.63  \\ 
%        Model Ensemble 1 &50.01&49.23 \\ 
%        Model Ensemble 2 &51.10&49.56  \\ 
%        \bottomrule
%    \end{tabular}
%    \caption{Table Example with 5 Columns and 6 Rows}
%    \label{tab:example}
%\end{table}

For categorical SER, we employ the state-of-the-art Emotion2Vec pipeline~\cite{emotion2vec} with model ensembling for emotion labeling. It classifies emotions into nine categories: \textit{angry, disgusted, fearful, happy, neutral, other, sad, surprised}, and \textit{unknown}. The pipeline has shown strong performance across multiple public datasets\footnote{https://github.com/ddlBoJack/emotion2vec}.  For dimensional SER, we utilize a model\footnote{Model: https://doi.org/10.5281/zenodo.6221127}~\cite{wagner2023dawn} that was created by fine-tuning the pre-trained wav2vec2-large-robust model\footnote{https://huggingface.co/facebook/wav2vec2-large-robust} on MSP-Podcast (v1.7)~\cite{Lotfian_2019_3}. 
To map categorical emotions  to dimensional emotion, we group the seven categories from Emotion2Vec (excluding the ``other'' and ``unknown'') into four sentiment classes: \textit{positive} $Y_\text{pos}$, \textit{negative} $Y_\text{neg}$, \textit{neutral} $Y_\text{neu}$, and \textit{ambiguous} $Y_\text{amb}$:
\begin{align}
Y_\text{pos}&=\{\text{\textit{happy}}\} \!
\quad
Y_\text{neu}=\{\text{\textit{neutral}}\}\!
\quad
Y_\text{amb}=\{\text{\textit{surprised}}\}\nonumber \\
Y_\text{neg}&=\{\text{\textit{angry, disgusted, fearful, sad}}\}%\nonumber
\end{align}
%
%\begin{align}
%    Y_\text{positive} &= \{\text{\textit{happy}}\} \\
%    Y_\text{neutral} &= \{\text{\textit{neutral}}\} \\
%    Y_\text{negative} &= \{\text{\textit{angry, disgusted, fearful, sad}}\} \\
%    Y_\text{ambiguous} &= \{\text{\textit{surprised}}\}
%\end{align}
%
To be noted, these mappings are not exhaustive, and there is potential to explore additional emotion categories and dimensions. In addition, we also annotated speaker gender to include in our QA generation task using WavLM-ECAPA\footnote{https://github.com/wenet-e2e/wespeaker} model \cite{chen2022wavlm, ECAPA} finetuned on the VoxCeleb2 dataset~\cite{chung2018voxceleb2}.

\subsubsection{Data condensation}

\renewcommand{\algorithmicrequire}{\textbf{Input:}}
\renewcommand{\algorithmicensure}{\textbf{Output:}}
\begin{algorithm}[t]
\setlength{\textfloatsep}{6pt}   % Reduce space above/below floats
\setlength{\floatsep}{6pt}       % Space between floats
\setlength{\intextsep}{6pt}      % Space between in-text figures and text
\caption{Filters for Data Condensation}
\label{alg:data-condensation}
\begin{algorithmic}[1]
\Require 
  $\mathcal{S} = \{\mathbf{s}_{1},\ldots,\mathbf{s}_{N}\}$: $N$ speech samples \\
  $\mathcal{C} = \{\mathcal{C}_{1},\ldots,\mathcal{C}_{N}\}$, $\mathcal{D} = \{\mathcal{D}_{1},\ldots,\mathcal{D}_{N}\}$ \\
  \quad where $\mathcal{C}_{i} = \{c_{i,1},\ldots,c_{i,M_i}\}$ \textit{(categorical labels)} \\
  \quad and $\mathcal{D}_{i} = \{d_{i,1},\ldots,d_{i,M_i}\}$ \textit{(dimensional values)} \\
  $\mathcal{Z}$: set of emotion categories \\
  $X_E, X_O$: filter conditions (consistency, occurrence)
\Ensure 
  $\widetilde{\mathcal{S}} \subseteq \mathcal{S}$: condensed set with reliable labels

\State $\widetilde{\mathcal{S}} \leftarrow \emptyset$
\For{$i \gets 1$ to $N$}
    \For{$j \gets 1$ to $M_i$}
        \Comment{1) Sub-segment Consistency}
        \If{($c_{i,j}, d_{i,j}$) is consistent w.r.t. $X_E$}
            \State  $c_{i,j} \gets c_{i,j}$  
        \Else
            \State $c_{i,j} \gets \text{``unknown''}$
        \EndIf
    \EndFor
    \ForAll{$Z_{k} \in \mathcal{Z}$}
        \Comment{2) Label Assignment}
        \If{$\left|\{c_{i,j} == Z_k \}\right| \in X_O(Z_k)$}
            \State \textbf{label} $\mathbf{s}_{i}$ as $Z_{k}$
            \State $\widetilde{\mathcal{S}} \leftarrow \widetilde{\mathcal{S}} \cup \{\mathbf{s}_{i}\}$
            %\State \textbf{break}
        \EndIf
    \EndFor
\EndFor
\State \Return $\widetilde{\mathcal{S}}$
\end{algorithmic}
\end{algorithm}

To ensure meaningful CPQA pairs with sufficient reasoning context, we first filter the dataset based on audio length, discarding speech segments shorter than a pre-determined threshold $\tau$. Next, we condense speech data using automatically estimated discrete emotion classes and valence values in emotional dimensions. The condensation process involves SER consistency filtering and occurrence filtering (see Algorithm~\ref{alg:data-condensation}), applied under the following conditions:
%\begin{enumerate}
\begin{itemize} 
\item SER consistency condition $X_E$: Ensures consistency between sentiment class mapped from discrete emotion categories and valence values. A sub-segment $s_{i,j}$ satisfies $X_E$, a set of the following:
\begin{align}
   c_{i,j} &\in Y_{\text{pos}} \quad \land \quad v_{i,j} \geq v_\text{pos,min} \\
  c_{i,j} &\in Y_{\text{neg}} \quad \land \quad v_{i,j} \leq v_\text{neg,max} \\
  c_{i,j} &\in Y_{\text{neu}} \quad \land \quad v_\text{neu,min} \leq v_{i,j} \leq v_\text{neu,max} \\
  c_{i,j} &\in Y_{\text{amb}} 
\end{align}
Here, $c_{i,j}$ and $v_{i,j}$ denote the estimated emotion category and valence for sub-segment $s_{i,j}$. Thresholds $\{v_\text{neu,min}, v_\text{neu,max},v_\text{pos,min},v_\text{neg,max} \}$ are set empirically or optimized based on labeled datasets. 

\item Occurrence condition $X_\text{O}$: Filters a subset that has greater confidence in speech $s_i$ belongs to a specific emotion category by applying a threshold $\alpha$ on the number of labels of that emotion within $s_i$. Due to the class imbalance, $\alpha$ varies across emotions. When $\alpha=1$, the filter is disabled, retaining all segments.
\end{itemize}
%\end{enumerate}

%\begin{figure}[t]
%  \centering
%  \includegraphics[width=\linewidth]{dimension.PNG}
%  \caption{Heatmap of UWA(\%) performance on the SG TV/movie dataset for varying factors $x$ and $y$. The triangle marks the best performance.}
%  \label{fig:dimensionSER}
%\end{figure}

\begin{figure}[t]
  \centering
  \includegraphics[width=0.85\linewidth]{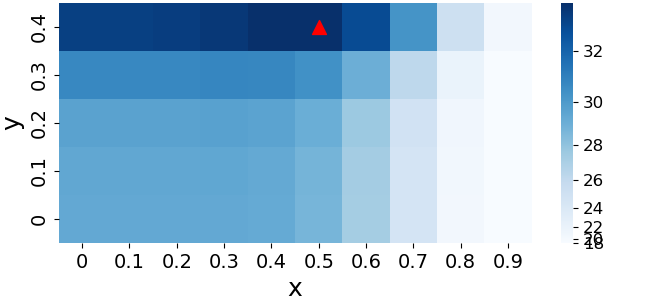}
  \caption{Heatmap of UWA(\%) performance on the SG TV/movie dataset for varying factors $x$ and $y$. The triangle marks the best performance.}
  \label{fig:dimensionSER}
\end{figure}

\subsection{Automated speech QA generation}
\begin{figure}[h]
    \raggedright  
    %\hspace{-24cm} 
    %\centering
    \begin{tcolorbox}[colframe=black, colback=gray!10, arc=2mm, boxrule=0.5pt, width=0.47\textwidth]
        \textbf{Prompt for Generating QA Pairs from Audio Clips}

        Generate diverse paralinguistic, content-based, and contextual reasoning QA pairs from a given audio clip. These QA pairs will train multimodal LLMs to reason from audio and text without relying on metadata.

        \textbf{Guidelines:}
        \begin{enumerate}
            \item \textbf{Focus Areas:}
            \begin{itemize}
                \item Explore speaker attributes such as emotion, gender, speaker transitions, and emotional reasoning.
                \item Include content-based questions, balancing simple and complex reasoning.
            \end{itemize}

            \item \textbf{Multimodal Integration:}
            \begin{itemize}
                \item Use both emotion labels and transcriptions, refining questions when labels are inaccurate.
                \item Ensure reliance on audio features without assuming metadata availability.
            \end{itemize}

            \item \textbf{Diversity and Depth:}
            \begin{itemize}
                \item Avoid repetitive or overly narrow questions.
                \item Encourage deeper multimodal reasoning.
            \end{itemize}

            \item \textbf{Output Format:}
            \begin{itemize}
                \item Format each QA pair using \texttt{Q:} and \texttt{A:} tags.
            \end{itemize}
        \end{enumerate}

        \textbf{Inputs:} Utterance: \texttt{`\{utterance\}`}, Word-level paralinguistics data: \texttt{\{word\_level\_data\}}.
    \end{tcolorbox}
    \vspace{-2pt}
   \caption{Prompt for Generating QA Pairs from Audio Clips}
    \label{fig:qa_prompt}
\end{figure}

To use text LLMs for QA generation, we need to provide time-aligned speech transcript and paralinguistic metadata information. We utilized WhisperX to generate word-level alignments~\cite{Bain2023WhisperXTS}. To align the emotion and gender metadata with the word-level transcript, each word is matched to the corresponding paralinguistic segment based on its timestamp. The process includes
temporal overlap-based matching: Each word's start and end time is compared with the time intervals of emotion and gender annotations. If a word falls within or overlaps with a paralinguistic segment, it inherits the corresponding emotion label and gender label.
%Handling Missing or Overlapping Segments: If multiple emotion labels apply to a word, priority is given based on segment duration from the emotion classifier.
This alignment ensures that each word is enriched with affective and speaker identity cues, making it valuable for emotion-aware speech processing, speaker profiling, and multimodal sentiment analysis. After alignment, we used specific prompt shown below to generate QA pair. The QA generator LLM used in our evaluation is  GPT4o API (2024-07-01-preview version)\footnote{https://learn.microsoft.com/en-us/azure/ai-services/openai/} from Azure.
We use the prompt in Figure~\ref{fig:qa_prompt} for the QA generation.
\section{Evaluation dataset creation}
We construct a dataset by applying our proposed data creation framework to speech data collected from top Singaporean YouTube channels.  
\subsection {Preliminary study for framework parameters} 
To set framework parameters, we conduct a preliminary study on an internal emotion dataset, SG TV/Movie dataset, comprising 117k speech segments (120 hours) from Singaporean TV shows and movies, primarily in English with some Mandarin. The emotion labels are annotated by human annotators. 

In pre-processing, we set $t = 2$ sec and $\Delta t = 1$ sec for automatic emotion labeling by SER. Additionally, gender labels were obtained using $t = 2$ sec and $\Delta t = 0.5$ sec in the same manner stated in Section~\ref{sssec:pre-processing}. %generated by an in-house WavLM-ECAPA model trained on the VoxCeleb2 dataset. 
For emotion labeling, we evaluated the Emotion2vec models on the SG TV/Movie dataset, excluding \textit{embarrassment}, \textit{sarcasm}, and \textit{worry} from the dataset's original ten emotion classes for compatibility. Zero-shot inference showed that the \textit{emotion2vec base} model exhibited the lowest performance among four individual models.
%Fig.~\ref{fig:sgtv} compares the performance in different Emotion2Vec+ models.  The \textit{emotion2vec base} model performs the worst, 
The best performance was achieved using an ensemble of three \textit{emotion2vec+} models~\cite{emotion2vec} by averaging their posteriors. It achieved accuracy of 51.10\%, unweighted accuracy (UWA) of 29.25\%, and F1-score of 49.56\%. Thus, it was selected for categorical SER.

 We apply length filtering with a minimum duration of $\tau = 30$ sec. For sub-segment SER consistency filtering, we explore UWA across different valence thresholds. Unlike accuracy, UWA mitigates selection bias toward dominant categories, addressing the varying difficulty of emotion categories in SER. 
 Thresholds in $X_E$ are set as $v_{\text{pos,min}} =x, v_{\text{neg, max}} =1-x, v_{\text{neu, min}}=y$, and $v_{\text{neu,max}} =1-y$.  Figure~\ref{fig:dimensionSER} shows the highest UWA 33.65\% is achieved at $x=0.5$ and $y = 0.4$ on the SG TV/movie dataset. For occurrence filtering, we set $\alpha = [10, 10, 4, 4, 2, 3]$ for the six non-neutral emotion categories: \textit{angry, disgusted, fearful, happy, sad}, and \textit{surprised}, to ensure sufficient and balanced samples are retained for each category in the condensed dataset.

\subsection{CPQA Evaluation dataset}
%Statistics, human annotation guideline...
%add in supplementary material: 2 samples, emotion label, question, answer. 

We randomly selected 80 samples for each emotion category from the condensed data to construct a balanced evaluation dataset. This resulted in a total of 480 audio samples, each ranging from 30 to 60 sec, corresponding to approximately 6.5 hours of data. From these samples, we generated 2,647 CPQA pairs with the ChatGPT LLM. For comparison, two human annotators created a CPQA set for the same data, with the objective of involving both paralinguistics and content for reasoning.

Table~\ref{tab:qa_stats} shows statistics of the two QA sets.  Questions about emotion include the speaker's emotional state or feelings, while questions about the speaker focus on attributes like gender and number.  Questions regarding contextual paralinguistic reasoning explore the underlying causes of a speaker's emotions and feelings. Other questions include content, topics, relationship between speakers, etc. During a manual review of ChatGPT QA, we observed repetitive questions, for example, multiple variations asking about the reasons behind emotions or changes in emotional state to increase diversity. We also encountered some irrelevant questions that assumed the existence of a text transcript (e.g., “What is the content in the audio from the text transcript?”). These questions were removed by post-filtering for keywords such as ``text'' or ``transcript''.

Emotion-related question-answer pairs are notably more frequent, while contextual reasoning and content-related questions are slightly less common in the model-generated set compared to human-generated one. As a result, out of 850 QA pairs, many questions belong to the CPQA section or associated with other attributes, such as gender and speaker identity. The LLM also generates general questions about content and topics, etc. Since the human annotators are explicitly instructed to focus on paralinguistic aspects, only 20 non-paralinguistic questions in their set, compared to 179 when using ChatGPT. Although this diversity in QA is valuable, the LLM generated fewer speaker-related questions due to its inability to process audio with multiple speakers of the same gender. To address this, future work will incorporate speaker diarization.

\begin{table}[t]
\centering
\caption{Comparison of Human- and ChatGPT-generated QA.}
\setlength{\tabcolsep}{1pt}
\label{tab:qa_stats}
\begin{tabular}{lrr}
\toprule
{} &  Human &  LLM \\
\midrule
Total QA Pairs                     &      2184 &        2647 \\
Questions about Emotion    &       150 &        850 \\
Questions about Speakers &       484 &         228 \\
%Questions about Speaker Gender     &       326 &         141 \\
%Questions about Reason for Emotion &       626 &         728 \\
%Short Answers (1-2 words)          &        26 &           0 \\
%Mentions of Multiple Speakers      &       122 &         326 \\
%Emotion Transitions                &        26 &         160 \\
Contextual Paralinguistic Reasoning Questions     &       1530 &         1390 \\
%Questions about speakers & 64 & 0 &\\
%Questions about relationship between speakers & 140 & 253\\
Others (content, topics, etc.) & 20 & 179\\
\bottomrule
\end{tabular}
\end{table}

\section{Evaluation}
\begin{figure}[t]
  \centering
  \includegraphics[width=0.9\linewidth]{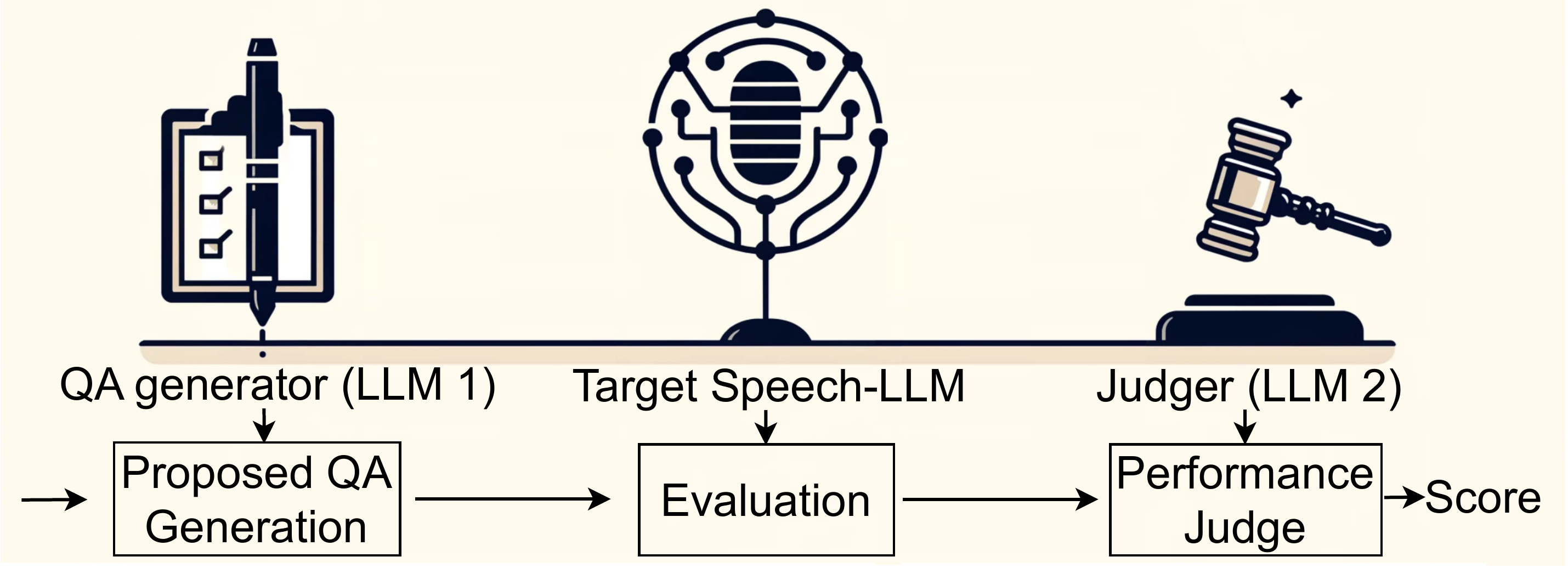}
  \caption{The illustration of the evaluation pipeline.}
  \label{fig:eval}
\end{figure}

%qq
 We validate the ChatGPT-generated CPQA set by evaluating Qwen2-Audio-7B-Instruct\footnote{https://huggingface.co/Qwen/Qwen2-Audio-7B-Instruct} speech-LLM since it is the best performing open source model as shown in large scale MMAU evaluation \cite{MMAU}. 
 To interpret the performance, we use Llama-3-70B-Instruct-AWQ\footnote{https://huggingface.co/casperhansen/llama-3-70b-instruct-awq} and ChatGPT-4o (2025-01-29 version)\footnote{https://help.openai.com/en/articles/9624314-model-release-notes\#} as two judger LLMs.  Qwen2~\cite{qwen2} and Llama-3~\cite{llama3} are deployed using a single NVIDIA H100 GPU (80GB). Notably, the QA generation and performance judgers use text-LLMs, while the evaluation uses a speech-LLM (see Figure~\ref{fig:eval}).
%\begin{figure}[t]
%  \centering
%  \includegraphics[width=\linewidth]{evaluation.drawio.png}
%  \caption{Evaluation}
%  \label{fig:eval}
%\end{figure}

%\begin{figure}[t]
%  \centering
%  \includegraphics[width=0.85\linewidth]{evaluation.pdf}
%  \caption{Evaluation}
%  \label{fig:eval2}
%\end{figure}

%\vspace{-3pt}

%\subsection{Settings}
% Tianchi
%The target Speech-LLM being evaluated is Qwen2-Audio-7B-Instruct\footnote{https://huggingface.co/Qwen/Qwen2-Audio-7B-Instruct}, while the judging LLMs are Llama-3-70B-Instruct-AWQ\footnote{https://huggingface.co/casperhansen/llama-3-70b-instruct-awq} and ChatGPT-4o, with the version released on January 29, 2025\footnote{https://help.openai.com/en/articles/9624314-model-release-notes\#}. Both Qwen2~\cite{qwen2audio} and Llama-3~\cite{llama3} are each deployed using a single NVIDIA H100 GPU with 80GB of memory.
%The number of manually annotated paralinguistic-related question-answer (QA) pairs is 2,184, while the number of automatically generated pairs using ChatGPT is 3,099.
Two evaluation prompts are employed for judgment. one from AudioBench~\cite{audiobench} that focuses on content accuracy and relevance and a refined version (Prompt 2). 
Prompt 2 incorporates both speech content and paralinguistic information and removes the penalty for brevity. Since Qwen2 processes audios up to 30 sec, we evaluate both the first and the last 30 sec of each segment and use the higher score for the QA pair. The final evaluation performance is the average score across all QAs. 
%, making it more suitable for paralinguistic evaluation tasks.
%assigning high scores to responses that are correct but concise, 

%We report the average score of these two segments, as well as the higher score between the two.

% Judge: Llama 70B
%\subsection{Results and analysis}
\begin{table}[t]
    \centering
    \caption{Zero-shot evaluation of CPQA sets. }
    \begin{tabular}{ccc c}
        \hline
        Judge& Prompt & LLM QA & Human QA \\ \hline
        \multirow{2}{*}{Llama 70B} & Prompt 1 & 53.86 &  52.29 \\ %\cline{2-4}
        & Prompt 2 &  56.82 & 54.33 \\ \hline
        \multirow{2}{*}{ChatGPT} & Prompt 1 &59.64  & 56.59 \\ %\cline{2-4}
        & Prompt 2 & 60.28 &  59.46 \\ \hline
    \end{tabular}
    \label{tab:eval}
\end{table}

%\subsection{Results and analysis}
%\begin{table}[t]
%    \centering
%    \begin{tabular}{|c|c|c|c|}
%        \hline
 %       Judge& Prompt & \textbf{LLM QA} & \textbf{Human QA} \\ \hline
%        \multirow{2}{*}{llama 70B} & Prompt 1 & 54.35 &  52.29 \\ \cline{2-4}
%        & Prompt 2 &  56.82 & 54.33 \\ \hline
%        \multirow{2}{*}{ChatGPT} & Prompt 1 &59.64  & 56.59 \\ \cline{2-4}
%        & Prompt 2 & 60.28 &  59.46 \\ \hline
%    \end{tabular}
%    \caption{Evaluation results on Qwen2-Audio-7B-Instruct model on the same speech %samples with both QA sets generated by LLM (ChatGPT) and human.The results by two judges: Llama-3-70B and chatGPT. }
 %   \label{tab:eval}
%\end{table}
The results in Table~\ref{tab:eval} indicate that the Qwen2 model demonstrates comparable performance on both LLM- and human-generated QA sets. This finding provides evidence, to some extent, that LLM-generated QA can serve as a viable tool for evaluating speech-LLMs. Furthermore, Prompt 2, which has been refined to align more closely with the contextual paralinguistic task, is expected to enhance the model's ability to assess evaluation performance more effectively. The consistent improvement is observed for both QA sets when using Prompt 2. Additionally, the observed correlation between the two QA sets further supports the validity of LLM-generated CPQA as a reasonable approach to evaluate speech-LLMs.

\section{Summary}
We propose a novel framework for generating dataset with contextual paralinguistic QA (CPQA) pairs from in-the-wild speech data, addressing the scarcity of data available for developing empathetic speech-LLMs. Our framework consists of pseudo paralinguistic label-based data condensation and LLM-based CPQA generation. We release a benchmark dataset comprising 480 speech samples. Evaluation using the Qwen2-Audio-7B-Instruct model, alongside comparisons with a human-generated set, demonstrates both the effectiveness of our dataset and the limitations of current speech-LLMs in empathetic reasoning. Our dataset provides a benchmark for evaluating speech-LLMs' contextual paralinguistic reasoning capabilities. Future work will explore more comprehensive assessments of paralinguistic contextual reasoning in speech-LLMs.

%We validate the effectiveness of our approach through an evaluation of the Qwen2-Audio-7B-Instruct speech-LLM on the generated dataset, comparing its performance against human-generated CPQA pairs. The results reveal the speech-LLM's limitations in handling empathetic reasoning tasks, highlighting the need for such datasets and more robust models.  
%\section{Acknowledgements}
%The authors would like to thank Jinyang Wu, 
%Nattadaporn Lertcheva, and Nabilah Binte Md Johan (\textcolor{red}{Ask Jinyang who else helped him  determin to keep or discard the channels?}) for their contribution to data collection and annotation. 

\section{Acknowledgement}
This research/project is supported by the National Research Foundation, Singapore, under its National Large Language Models Funding Initiative. Any opinions, findings, conclusions or recommendations expressed in this material are those of the author(s) and do not reflect the views of the National Research Foundation, Singapore.

\bibliographystyle{IEEEtran}
\bibliography{mybib}

\end{document}